%% file: main.tex
\documentclass[10pt,twocolumn,letterpaper]{article}

\usepackage[pagenumbers]{iccv} 


\definecolor{cvprblue}{rgb}{0.21,0.49,0.74}
\usepackage[pagebackref,breaklinks,colorlinks,allcolors=cvprblue]{hyperref}

\usepackage{color,soul}
\usepackage{multirow}
\usepackage{graphics}
\usepackage{xspace}

\input{s00_Commands}

\input{s01_Header}

\begin{document}

 \maketitle
 \thispagestyle{empty}
 \pagestyle{empty}

\input{s02_Abstract}
 \input{s10_Intro}

 \input{s20_Related_Work}

\input{s30_Dataset}
 \input{s40_Methodology}
 \input{s50_Results}

 \input{s60_Conclusions}

  
 \bibliographystyle{ieeenat_fullname}
 \bibliography{s99_References}

\end{document}


\maketitle
 \thispagestyle{empty}
 \pagestyle{empty}

\setcounter{figure}{4}
\setcounter{table}{4}
\setcounter{section}{2}

\section{The \datasetName Dataset}
\label{sec:dataset}

\begin{figure*}[ht]
  \centering
  \includegraphics[width=\textwidth]{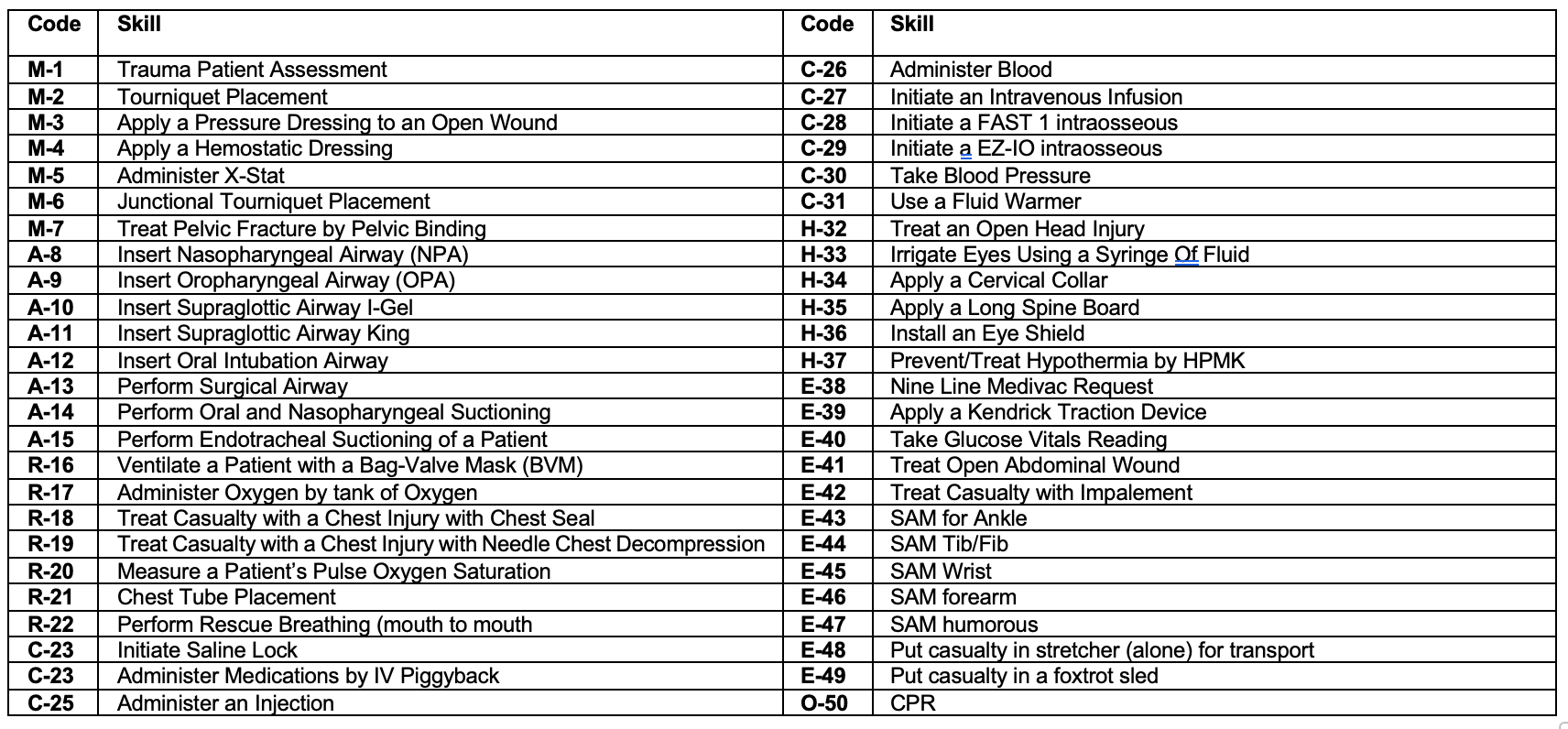} 
  \caption{  List of all tasks record with description and letter-number code related to MARCH algorithm.  }
  \label{fig:all_skills}
\end{figure*}

\begin{figure*}[ht]
  \centering
  \includegraphics[scale=0.6]{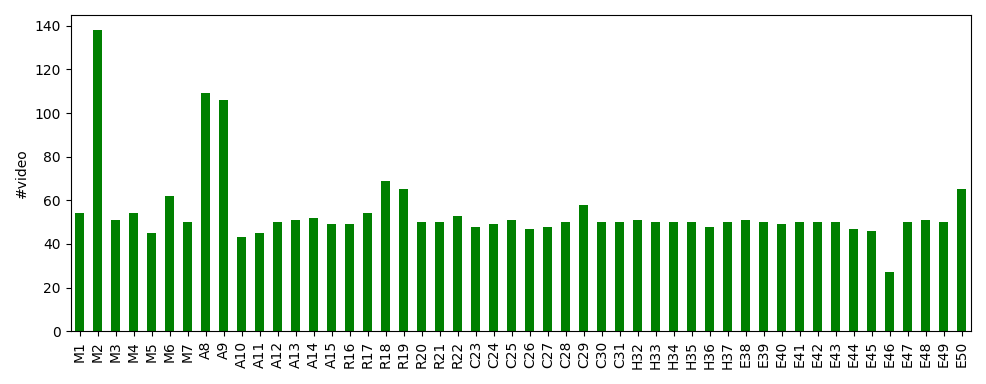} 
  \caption{  Distribution of videos per tasks.  }
  \label{fig:videos_per_task}
\end{figure*}

\begin{figure*}
  \begin{subfigure}{0.5\textwidth}
    \centering
    \includegraphics[width=\textwidth]{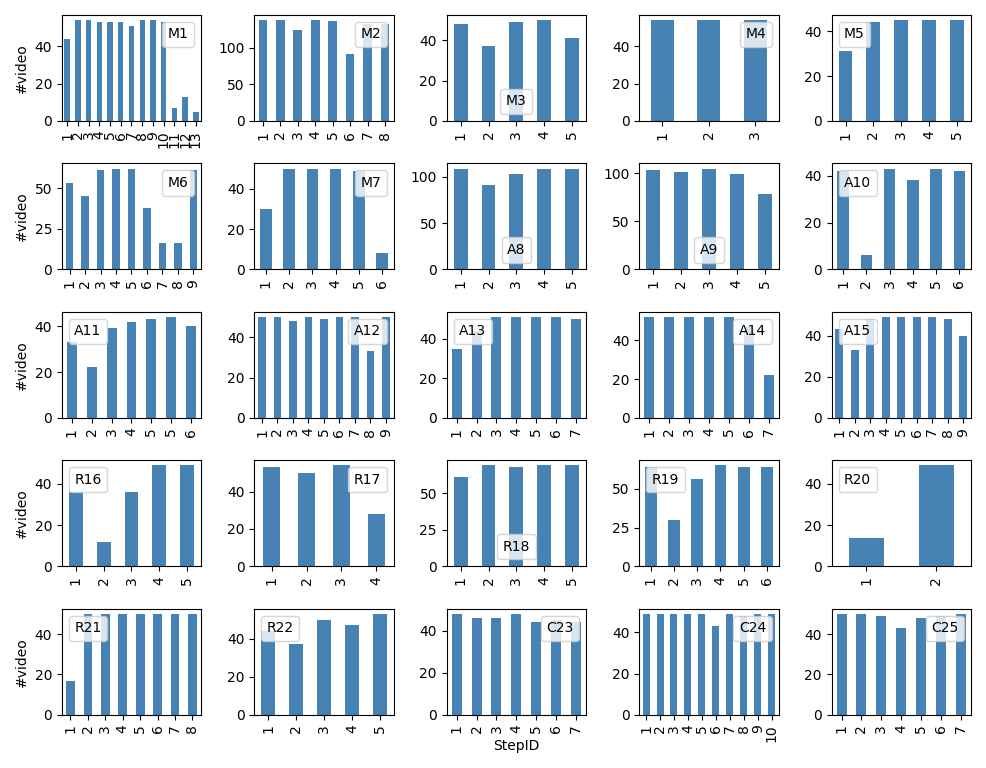}   
  \end{subfigure}
  \hfill
  \begin{subfigure}{0.5\textwidth}
    \centering
    \includegraphics[width=\textwidth]{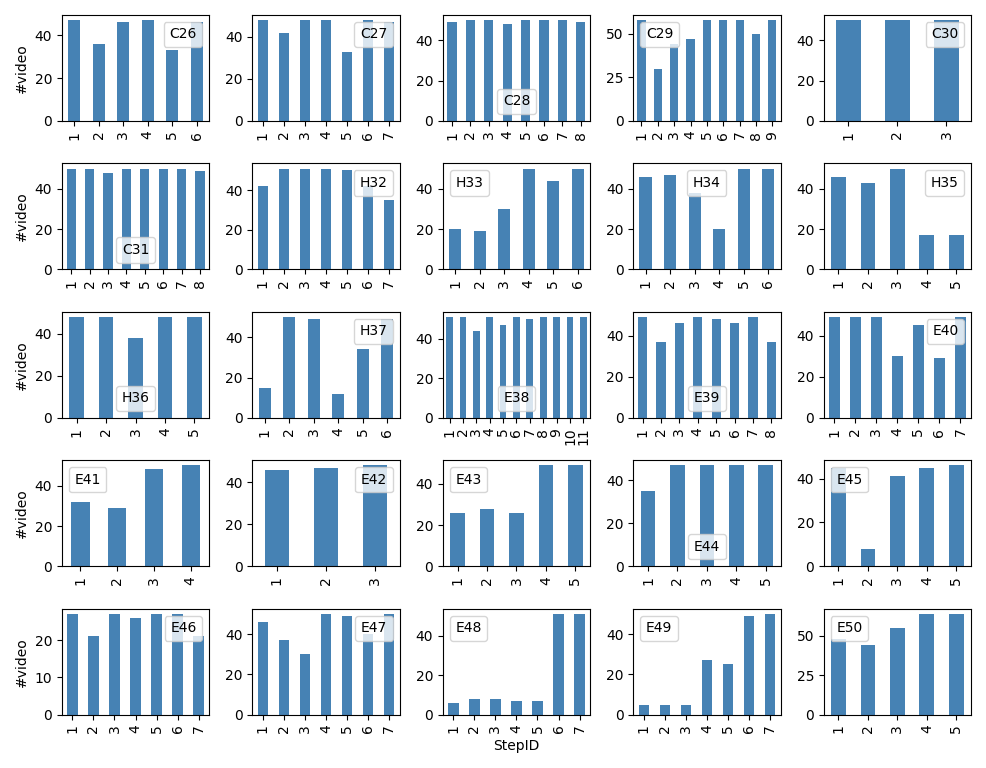}   
  \end{subfigure}
  \caption{  Distribution of videos per steps.  }
  \label{fig:videos_per_step}
\end{figure*}

\begin{figure*}
  \begin{subfigure}{0.5\textwidth}
    \centering
    \includegraphics[width=\textwidth]{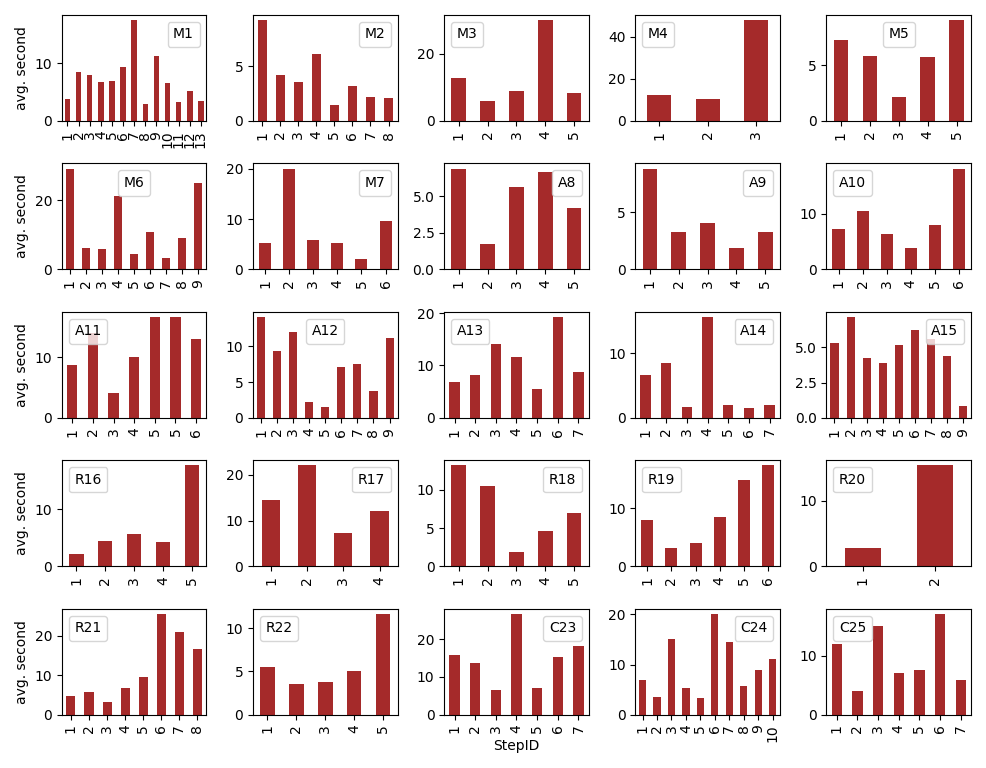}    
  \end{subfigure}
  \hfill
  \begin{subfigure}{0.5\textwidth}
    \centering
    \includegraphics[width=\textwidth]{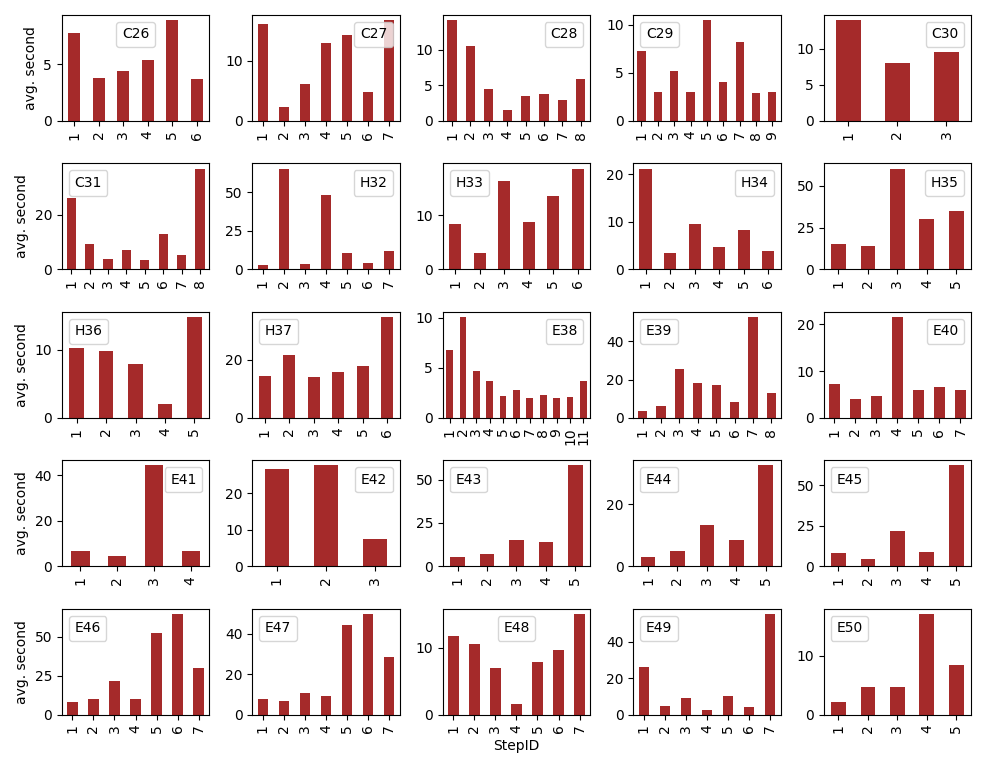}    
  \end{subfigure}
  \caption{ Average step size in seconds.  }
  \label{fig:step_avg}
\end{figure*}

\datasetName consists of \numTotalSkills tasks listed in Figure~\ref{fig:all_skills}.
Each task has a letter-number code, where the letter corresponds to the MARCH algorithm for treatment: massive bleeding, airway, respiration, circulation, head/hypothermia; and the number is a sequential identifier of tasks from each treatment.
Figures~\ref{fig:videos_per_task} and~\ref{fig:videos_per_step} present the video distribution by task and step, respectively. 
Figure~\ref{fig:step_avg} presents the average size of the steps, highlighting one of the data challenges: shorter steps can be more difficult to predict than longer ones.

\section{Action Detection Benchmark}


\subsection{Recurrent and transformer approaches}

To deal with challenging scenarios with different environmental conditions, object occlusion, and overlap of task steps, one of the \ptg research teams developed the following approaches. All code will be available online on GitHub.
The next sections present details about data preprocessing, training process, hyperparameters, and basic evaluation of the related models presented in the main text.

\subsubsection{State machine} 
Beyond step prediction, we associated to the model outputs a state to each step with a simple rule-based state machine.
It creates an internal state vector with a size equal to the step number and initializes all positions with \emph{unobserved} state.
The state machine smooths each position of the last $p$ predicted probability vectors ($p = 3$) by applying the median.
After smoothing predictions, it normalizes the final vector to guarantee sum $= 1.0$.
If the step with the highest probability has a state different from \emph{current}, its state is updated to \emph{current} after all other \emph{current} states are updated to \emph{done}.
At this point, all unpredicted steps are \emph{unobserved}, all previously visited steps are \emph{done}, and only one step is \emph{current}.
The median sliding window over the predictions means the state machine is less influenced by subtle changes in the prediction and only updates the internal state if the model predictions are consistent over time.

\subsubsection{Video windowing}
\label{sec:nyu_windowg}

\begin{figure*}[!h]
  \centering
   \includegraphics[scale=1.0]{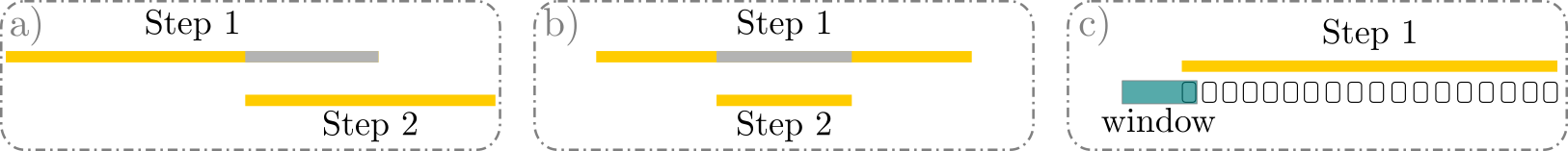} 
\caption{ 
    Before sliding a window, the RNN and Swin-T processes deal with basic overlap scenarios, a) and b). All Step 1 shaded parts are discarded during the training process.
    In the beginning, the last window frame is aligned with one of the first step frames as in c).  }
  \label{fig:nyu_step_overlap}   
\end{figure*}

For training purposes, before sliding a window over a video, our process removes any possible overlap between the steps annotations, and \autoref{fig:nyu_step_overlap} presents two basic overlap scenarios -- a) and b).
Step 1 shaded portion overlaps Step 2 and that shaded portion is removed from Step 1.
With this procedure, the model learns both steps separately and decides which step to focus on when overlapped.
Besides, any non-labeled video part is considered a no-step, i.e., the person in the video is not performing any important action for a task. 
After these two procedures, the step labels and the additional no-step label cover the video timeline.

Following this, the process considers each step a small video clip and slides a $k$-second window over it, with default $k = 2$, and overlap $=50\%$.
All the windows have the same step label of the processed video clip (step). 
As in~\autoref{fig:nyu_step_overlap}c, the process starts with the last window frame aligned with one of the $fs$ first-step frames, being $fs = 5$ -- the smallest step size in the dataset. 
The process randomly chooses one of the $fs$ first frames to start the process.

Beyond the window step label, we also consider a positional label defined as $[begin, end]$ to help model training, being $begin$ the relative distance between the first step frame and the last window frame, and $end = 1.0 - begin$.

To deal with a variety of video frame rates, the process loads the window frames considering $k$-seconds and loads the needed number of frames based on each video frame rate.
It works as an induced time augmentation process, showing models different frame rates.

\subsubsection{Image augmentation}
\label{sec:nyu_augmentation}

\begin{table*}[!h]
\centering
\caption{ Stacked augmentation techniques applied to each video to train RNN and Swin-T models. }
\label{tab:nyu_augmentation_params}
\begin{tabular}{clll}
\toprule
\multicolumn{2}{c}{technique}                                      &  application   & hyperparameter range \\
\midrule
\multicolumn{1}{l}{}         & horizontal flip                     &    50\%        &                      \\
\multicolumn{1}{l}{}         & perspective change                  &                &  start = 0, end = 0.9, inc = 0.025 \\
\midrule
\multirow{4}{*}{color alter} & brightness                          &                &  start = 0.7, end = 1.5, inc = 0.05 \\
                             & gamma                               &                &  start = 0.5, end = 2.0, inc = 0.05 \\
                             & hue                                 &                &  start = -0.05, end = 0.05, inc = 0.001 \\
                             & contrast, saturation, and sharpness &                &  start = 0.5, end = 1.5, inc = 0.05 \\
\midrule
\multicolumn{1}{l}{}         & black rectangular mask              &   50\%         & $[1/3, 1/4, 1/5]$ of the image \\
\multicolumn{1}{l}{}         & salt and pepper noise               &                & start = 900, end = 1000, inc = 10 \\
\bottomrule
\end{tabular}
\end{table*}

The augmentation process increases data variance by randomly choosing $80\%$ of the videos, applying a stack of augmentations to their frames, and replacing the original videos with the augmented ones.
Therefore, a model is trained with $80\%$ of augmented videos and $20\%$ of regular ones.

\autoref{tab:nyu_augmentation_params} lists the techniques considered and their hyper-parameter range.
Some techniques are applied only in an \emph{application} fraction of the video inputs.
In addition, a simple random walk procedure chooses the input values for each technique in the presented \emph{range}.
We chose these techniques to help the models learn in different camera positions, lighting, and noise.
Besides, the \emph{rectangular mask} proposed by~\cite{Gorpincenko:2023}, even if listed here as image augmentation, has strong capabilities to help the models to learn temporal patterns. 

\subsubsection{Training and evaluation}

\begin{table}[!h]
\centering
\caption{Professional and in-lab data used for RNN and Swin-T models training and evaluation.}
\label{tab:nyu_dataset}
\begin{tabular}{llll}
\toprule
task & professional & in-lab & total \\
\midrule
A8    & 109        & 69      & 178   \\
M2    & 138        & 76      & 204   \\
M3    & 50         & 50      & 100   \\
M4    & 54         & 49      & 103   \\
M5    & 45         & 47      & 92    \\
R16   & 49         & 50      & 99    \\
R18   & 69         & 83      & 152   \\
R19   & 65         & 39      & 104   \\
\midrule
total &579         & 463     & 1042 \\
\bottomrule
\end{tabular}
\end{table}

We trained four models (RNN and transformer models described in the main text) per task over the data presented in~\autoref{tab:nyu_dataset}.
However, only two models were deployed, the GRU and one Video Swin-transformer version, which performed the best.
The GRU is trained to detect the steps and the relative position of a window (see~\autoref{sec:nyu_windowg}).
Its loss function is the addition of a weighted cross-entropy and the mean square error (MSE).
The weights are applied to mitigate the imbalance of video window numbers per step.
The MSE is an auxiliary loss function that minimizes the error of window relative position and improves training stability and model accuracy.
On the other hand, all the Swin-transformers are trained only with the weighted cross entropy.

To train the models, we used batch size $= 8$ (limited by video card constraints), 25 epochs for GRU and 30 epochs for Swin-transformers, and other optimization hyper-parameters are presented in \autoref{tab:nyu_opt_params}.
We ran a grid search varying the optimizer (Adam and AdamW), the $learning rate = [0.01, 0.001, 0.0001]$, and weight-decay$= [0.00001, 0.0001, 0.001, 0.01]$ to find the best values for the GRU model and, for Swin-transformer models, we used the default values of its paper.

\begin{table*}[!h]
  \centering
  \caption{Optimization hyper-parameters of RNN and Swin-T models}
  \label{tab:nyu_opt_params}
  \begin{tabular}{lllllll}
  \toprule
  model & task         & opt.    & lr                     & lr scheduler      & warm-up         & weight-decay   \\
  \midrule
  \multirow{5}{*}{GRU} & A8,M4,R16,R19 & Adam    & 0.001                  &                   &                &      \\
  \cmidrule(lr){2-7}              
  & M2            & \multirow{4}{*}{AdamW}   & \multirow{4}{*}{0.001} &                   &                & 0.0001  \\
  & M3            &                          &                        &                   &                & 0.001   \\
  & M5            &                          &                        &                   &                & 0.00001 \\
  & R18           &                          &                        &                   &                & 0.01    \\
  \midrule
  \midrule  
  Swin & all           & AdamW   & 0.0001                 & cos$(T_{max}=5)$ & linear$(epoch=3)$ &  0.05           \\
  \bottomrule
  \end{tabular}
\end{table*}

For evaluation, the video windowing (see~\autoref{sec:nyu_windowg}) process starts by aligning the last window frame with the first video frame and slides the window throughout the video.
The window step is the label of the last window frame overlapped with the ground truth.
This process is closer to the real application of the model in video streaming.

Besides, we randomly select 10\% of each task in~\autoref{tab:nyu_dataset} for testing and the remaining for training, setting random seeds that could return videos recorded by professionals and in the BBN lab. 
With this approach, the models are trained and tested on both environments.
We considered evaluation metrics, such as precision, recall, f1-score, and weighted accuracy, using the same weights of the loss function.



\section{Experimental Results}

\subsection{Belief File Format}

\begin{figure*}[t!]
    \centering
    \begin{subfigure}[t]{1\textwidth}
        \centering
        \includegraphics[width=1\textwidth]{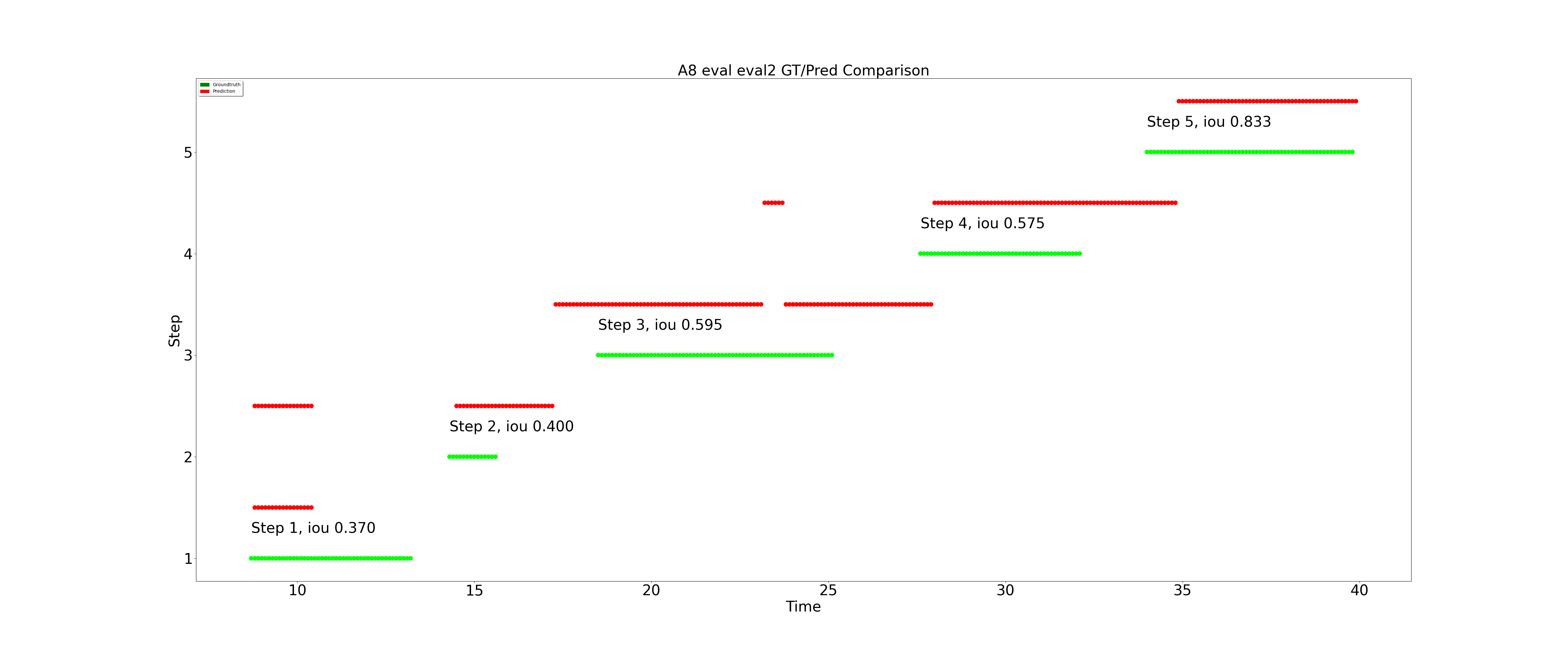}
    \end{subfigure}%

    \begin{subfigure}[t]{1\textwidth}
        \centering
        \includegraphics[width=1\textwidth]{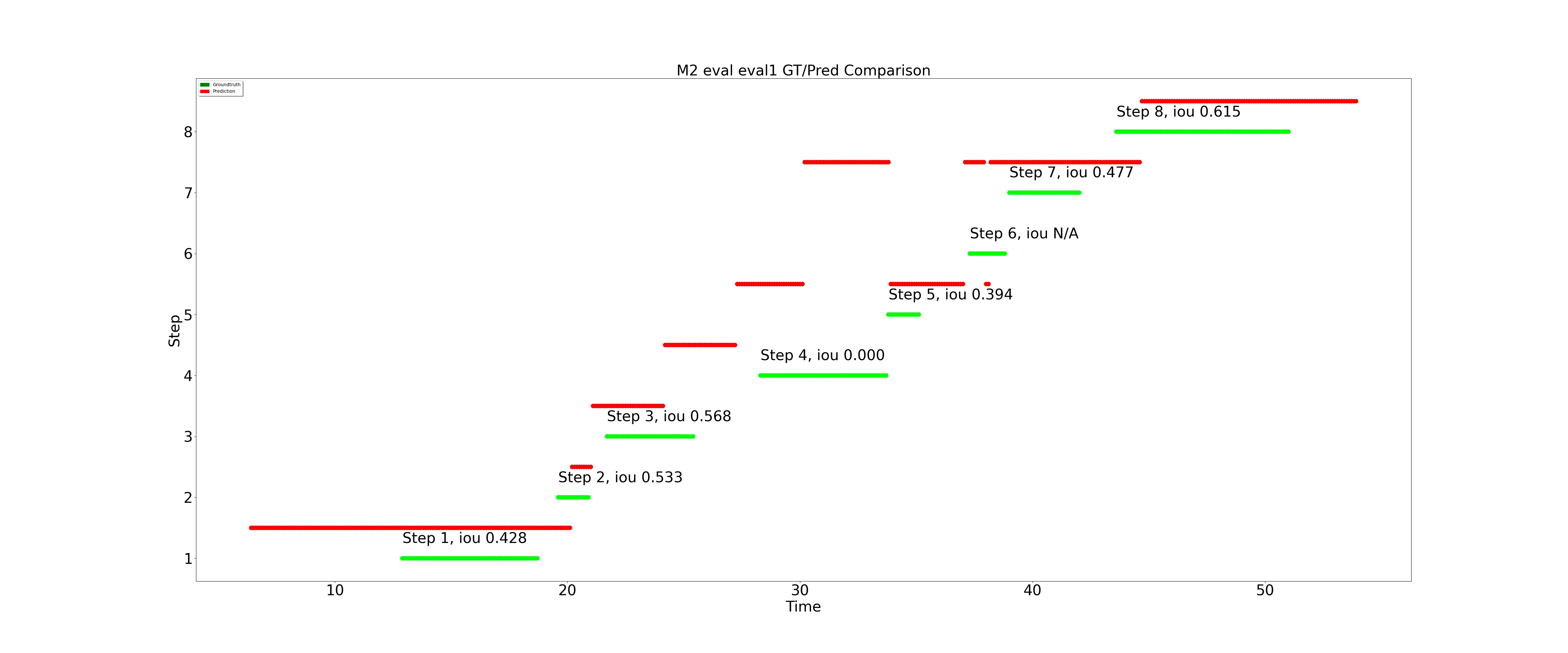}
    \end{subfigure}

    \begin{subfigure}[t]{1\textwidth}
        \centering
        \includegraphics[width=1\textwidth]{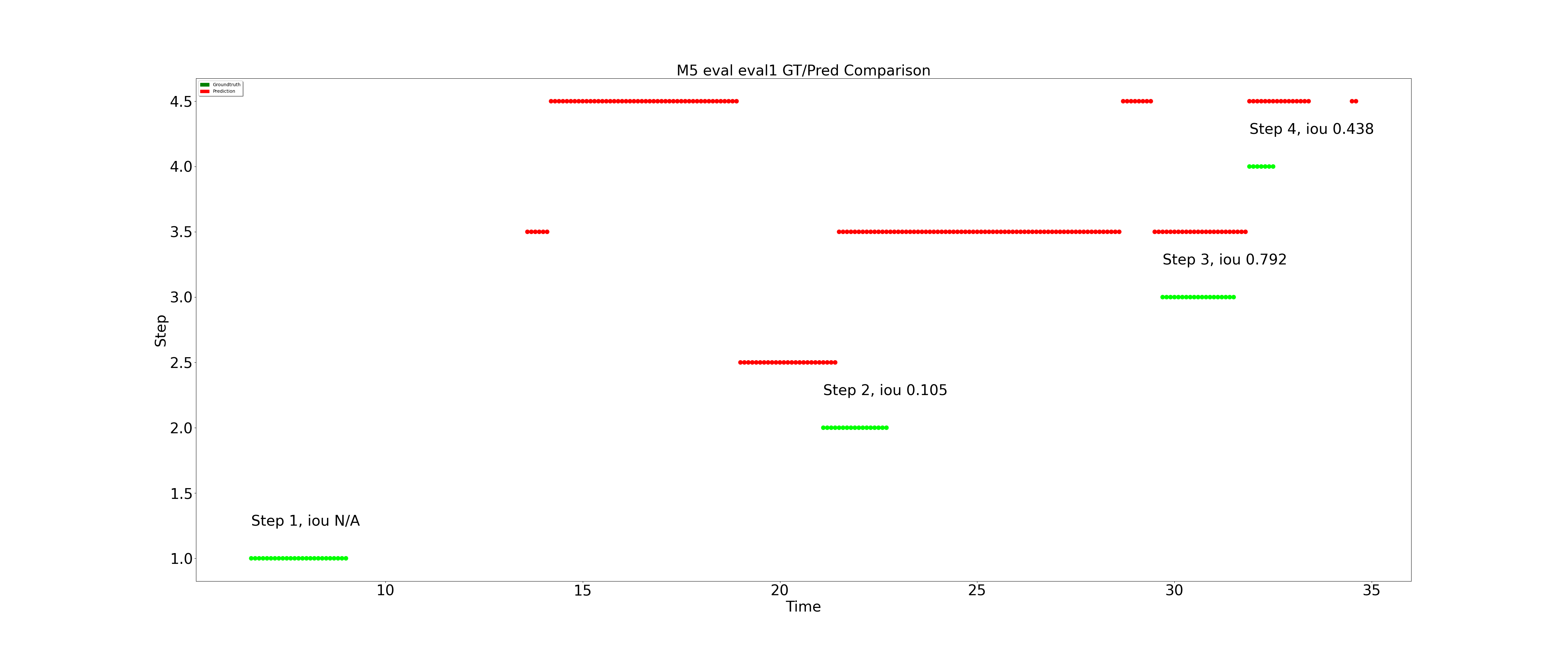}
    \end{subfigure}
    
    \caption{Example action segments from several different tasks.  Green lines represent ground truth data.  Red lines represent predicted values.  Predicted values come from the TAS method presented in the paper.}
    \label{fig:action_segments}
\end{figure*}

The evaluation process was heavily informed by the \ptg program, where the focus was on online real time operation.
The evaluation kit for the \datasetName dataset was adapted from what was being done under PTG.
This is where the belief file format comes from, which expects input in the form of a CSV file where each line contains:
\texttt{task\_code, task\_step\_num, step\_state, step\_state\_confidence, timestamp}, 
where \texttt{task\_code} is one of the aforementioned codes such as M2 or R18,
\texttt{task\_step\_num} is the step number, 
\texttt{step\_state} is one of unobserved, current, or done,
\texttt{step\_state\_confidence} is a confidence [0,1], 
and \texttt{timestamp} is a fractional second timestamp.
These belief files have one entry per task step per timestamp.
These belief files are processed in order to create action segments that are compared for evaluation.
First, the timesteps are binned across time into 0.1 second temporal bins.
When a step first enters the current step that is taken as the start of the action, and the first timestep the task step enters the done state is when the action is completed.
The evaluation kit handles all of this and outputs plots that show how ground truth action segments compare to the predicted action segments.
This process is also further documented in the evaluation kit documentation.
See Figure~\ref{fig:action_segments} for some examples of action segments from the methods presented in the paper.

 \bibliographystyle{ieeenat_fullname}
 \bibliography{s99_References}

%% file: s00_Commands.tex
\newcommand{\ptg}              {{PTG~}}

\newcommand{\magic}            {{MAGIC}\xspace}
\newcommand{\datasetName}      {{EgoMAGIC}\xspace}

\newcommand{\datasetUrl}{{\texttt{\url{https://github.com/BBN-VISUAL/egomagic_eval}}}\xspace}

\newcommand{\numVAMSInstructors}{{10}\xspace}
\newcommand{\numTotalSkills}   {{50}\xspace}
\newcommand{\numSelectedSkills}{{8}\xspace}
\newcommand{\numTotalVideos}   {{3355}\xspace}
\newcommand{\numProVideos}     {{2735}\xspace}
\newcommand{\numLabVideos}     {{620}\xspace}

\newcommand{\numFramesPro}     {{8.2M}\xspace}
\newcommand{\numFramesLab}     {{1.1}\xspace}
\newcommand{\numObjAnnos}      {{1,958,877}\xspace}
\newcommand{\numObjClasses}    {{124}\xspace}

\newcommand{\numProSSAnnos}    {{14,424}\xspace}
\newcommand{\numLabSSAnnos}    {{3,298}\xspace}
\newcommand{\numHandObjAnnos}  {{39,068}\xspace}

%% file: s01_Header.tex
\def\paperID{14553}
\def\confName{ICCV}
\def\confYear{2025}

\title{ \datasetName - An Egocentric Video Field Medicine Dataset for Training Perception Algorithms}

\author{Brian VanVoorst, Nicholas Walczak, Christopher Gilleo, Charles Meissner \\
RTX BBN Technologies \\
{\tt\small brian.vanvoorst@rtx.com, nicholas.walczak@rtx.com} \\
{\tt\small christopher.gilleo@rtx.com, charles.meissner@rtx.com}
\and 
Fabio Felix, Iran Roman, Bea Steers, Claudio Silva \\
New York University \\
{\tt\small ffd2011@nyu.edu, csilvia@nyu.edu, bsteers@nyu.edu, roman@nyu,edu}
\and 
Yuhan Shen, Zijia Lu, Shih-Po Lee, Ehsan Elhamifar \\
Northeastern University \\
{\tt\small eelhami@ccs.neu.edu, lu.zij@northeastern.edu,} \\
{\tt\small shen.yuh@northeastern.edu, lee.shih@northeastern.edu}
}

%% file: s02_Abstract.tex
\begin{abstract}

This paper introduces \textbf{\datasetName} (Medical Assistance, Guidance, Instruction, and Correction), an egocentric medical activity dataset collected as part of DARPA's Perceptually-enabled Task Guidance (PTG) program. 
This dataset comprises \textbf{\numTotalVideos} videos of \textbf{50 medical tasks}, with at least \textbf{50 labeled videos per task}. 
The primary objective of the \ptg program was to develop virtual assistants integrated into augmented reality headsets to assist users in performing complex tasks.
To encourage exploration and research using this dataset, the medical training data has been released along with an action detection challenge focused on eight medical tasks.
The majority of the videos were recorded using a head-mounted stereo camera with integrated audio.
From this dataset, \textbf{40 YOLO models} were trained using \textbf{1.95 million labels} to detect \textbf{124 medical objects}, providing a robust starting point for developers working on medical AI applications.
In addition to introducing the dataset, this paper presents \textbf{baseline results} on action detection for the eight selected medical tasks across three models, with the best-performing method achieving average mAP 0.526.
Although this paper primarily addresses action detection as the benchmark, the \datasetName dataset is equally suitable for action recognition, object identification and detection, error detection, and other challenging computer vision tasks. The dataset is accessible via zenodo.org (DOI: 10.5281/zenodo.19239154).

\end{abstract}

%% file: s10_Intro.tex
\section{Introduction}
\label{sec:introduction}

\input{figures/summary/summary_fig} 

The advances in machine learning video-related tasks are heavily related to the development and enhancement of powerful deep learning architectures, such as convolutional neural networks~\cite{Goodfellow:2016} and transformers~\cite{Vaswani:2017}, along with large-scale datasets, such as Youtube-8M~\cite{Haiha:2016}, Something-something~\cite{Goyal:2017}, Kinetics~\cite{Kay:2017}, EPIC Kitchens~\cite{epickitchens}, and Ego4D~\cite{Grauman:2022}.
Although these and other datasets help models achieve strong generalization, more specialized datasets are still needed in areas like medical procedures.
This paper introduces the \datasetName dataset (Medical Assistance, Guidance, Instruction, and Correction), an egocentric dataset designed to train deep learning models specifically for medicine tasks.
The dataset and benchmark models were developed as part of the Defense Advanced Research Projects Agency’s (DARPA) Perceptually-enabled Task Guidance (PTG) program~\cite{PTG_site} that aimed to create real-time augmented reality (AR) assistants that guide users through tasks. 
These assistants are intended to recognize medical procedures, automatically track completed steps, and provide guidance on the next step or corrections when steps are missed or performed incorrectly.

This scenario presents exciting opportunities for the deep learning community. 
To encourage research and exploration of the \datasetName dataset, the action detection component (identifying the start and stop time of individual steps) has been emphasized as a stand-alone challenge, which is the focus of this paper.
This challenge is structured similarly to other egocentric action detection datasets, such as EPIC Kitchens~\cite{epickitchens} and the Trauma THOMPSON challenge~\cite{thompsontrauma}. 

\noindent\textbf{Novelty \& Contributions.
}
Emergency response medical triage is inherently fast-paced and chaotic, occurring in unstructured environments with significant open-world variability \cite{phtls2019}. 
\ptg selected combat medicine as a domain due to its unique challenges and the significant value an AR assistant could offer to caregivers in high-stress emergencies.
In this scenario, the \datasetName dataset introduces several unique action detection challenges not found in other egocentric datasets, such as:

\begin{enumerate}
    \item Some steps within many tasks are extremely brief, lasting as little as a second or less (see Table~\ref{tab:dataset} and Figure~\ref{fig:step_avg}).
    \item Steps can frequently be performed concurrently, leading to overlapping activities.
    \item Some steps are optional, allowing for variability in step sequence and completion.
    \item Rapid egomotion is common throughout the activities, adding complexity to the visual data.
    \item Realistic clutter of objects and environment. Figure~\ref{fig:summary} top- and bottom-center present examples of a partial view of objects.
\end{enumerate}

While the primary focus is on action detection—identifying the start and stop times of steps in medical procedures—the dataset also supports other research challenges, including action recognition (identifying the action being performed within a given clip), and action anticipation (predicting future actions).
Furthermore, the dataset contains extensive object labels for medical equipment and supplies, making it valuable for object detection research.
The dataset includes \numTotalVideos videos covering \numTotalSkills medical tasks as highlighted in Figure~\ref{fig:summary}. For data, it provides:

\begin{itemize}
    \item \numFramesPro frames extracted from the left lens of the Zed stereo camera and \numFramesLab extracted from the HL2 main camera
    \item Over 1.95 million labeled objects across 124 object classes
    \item More than 17,000 task step annotations across 286 step classes
    \item Over 39,000 hand-object interactions
\end{itemize}

The dataset also provides 40 pre-trained YOLOv8~\cite{Redmon:2016} models to help developers with object detection and action recognition tasks.

In addition to introducing a new dataset, this paper presents the approaches and results of research teams that developed deep learning models for action detection in the \ptg scope.
The results work as a benchmark and challenge report over the \datasetName dataset.


%% file: figures/summary/summary_fig.tex
%
%
\begin{figure*}[ht]
  \centering
  \includegraphics[scale=1.0]{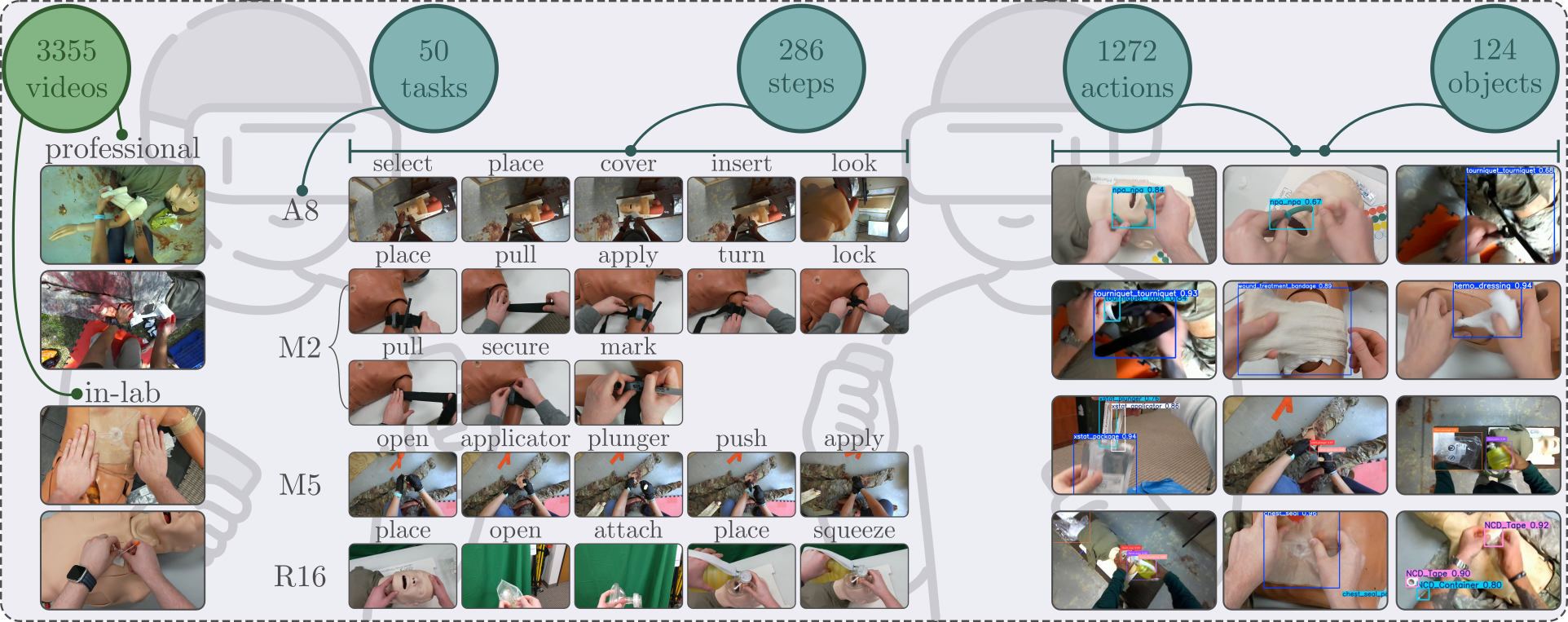}  

  \caption{ 
    Visual summary of the dataset showing the total number of videos and tasks, along with unique annotated labels for steps, actions, and objects.
    The images illustrate the recording environments (professional or in-lab), highlight the five most challenging tasks with keywords descriptions of their steps, and provide examples of objects related to specific actions. 
  }
  \label{fig:summary}
\end{figure*}
%

%% file: s20_Related_Work.tex
\input{figures/dataset_table}
\input{figures/dataset_step_avg}

\section{Related Datasets}
\label{sec:relatedWork}

\noindent\textbf{Egocentric Action Detection.} 
Identifying and detecting actions in egocentric videos is important for AR assistants to understand the environment and guide users through their tasks, even potentially catching user mistakes.
However, working with the egocentric view is challenging, as it includes camera motion, occlusions, and overlapping activities that can hinder the reliability of action detection.

There is a growing effort to provide datasets and benchmarks that help developers and researchers enhance deep learning models for AR assistants.
For example, Georgia Tech Egocentric Activity Dataset (GTEA)~\cite{Fathi:2011,Li:2015} and YouHome Activities-of-Daily-Living (ADL)~\cite{Pan:2022}, focused on daily activities and offered a glimpse into how people interact with objects in their immediate surroundings.
EPIC-Kitchens~\cite{epickitchens} and others built upon it, expanded the domain of egocentric vision to cooking activities in home kitchens, allowing researchers to examine fine-grained hand-object manipulations within a complex and cluttered environment.
Something-Something-v2 (SSv2)~\cite{Goyal:2017} shows users performing basic daily actions, the objects involved in each action, and labels with simple descriptions defined as caption templates.
Ego4D~\cite{Grauman:2022} has aimed to collect broader, real-world first-person experiences, for a wide range of daily tasks and interactions.

\noindent\textbf{Egocentric Medicine and Combat Medicine.}
Previous efforts focused on activity detection for creating video-based after action review (AAR) tools~\cite{vanvoorst2023automated,walczak2022coach}, in the context of medical training exercises. 
Most closely related to \datasetName, it is the Thompson Trauma dataset~\cite{thompsontrauma} and corresponding challenge which provides an egocentric video benchmark focused on medical procedures related to trauma care. It captures real-world conditions characterized by rapid movements, cluttered settings, and overlapping actions typical of emergency response scenarios. These complexities pose unique challenges for computer vision methods, demanding robust solutions capable of handling motion blur, occlusions, and the need for rapid, fine-grained recognition of critical medical steps. Furthermore, the domain of emergency care moves where accuracy and timeliness can have substantial real-world implications.

Our dataset expands on the Thompson Trauma dataset in several ways.
It increases breadth, moving from 5 to 50 procedures, and from 200 videos to \numTotalVideos.
Secondly, \datasetName provides three types of data annotations (task-step, hand-interaction, and object labels) described in Sec.~\ref{sec:dataset} and exemplified in Figure~\ref{fig:summary}. 

%% file: figures/dataset_table.tex
%
%
\begin{table*}[!ht]
\centering
 \scalebox{0.8}{
 \begin{tabular}{ lcccccc}
 \toprule
 \bf{Task  (8)}& \bf{Num Steps}    &  \bf{Num Videos} &  \bf{Avg Dur (sec)} &  \bf{Avg Step Dur (s)} &  \bf{Min Step Dur (s)} & \bf{\% of Steps with Overlaps}    \\
 \midrule

  NPA Tube [A8]&       5      &         109   &  45.12  &  5.12 & 0.81 &  0.0\%  \\
  Apply Tourniquet [M2]&       8      &         138   &  51.31  &  4.01 & 0.37 & 16.3\%  \\
  Pressure Dressing [M3]&       5      &          50   &  82.88  & 15.98 & 0.43 & 15.1\%  \\
  Wound Packing [M4]&       3      &          54   &  98.97  & 32.15 & 2.41 & 29.6\%  \\
  X-Stat [M5]&       5      &          45   &  41.37  &  7.55 & 0.65 &  0.0\%  \\
  Ventilate (BVM) [R16]&       5      &          49   &  41.54  &  6.85 & 0.43 &  1.1\%  \\
  Apply Chest Seal [R18]&       5      &          69   &  44.02  &  8.55 & 0.13 & 14.8\%  \\
  Needle Chest Decomp [R19]&    6      &          65   &  81.15  &  9.72 & 0.43 &  4.7\%  \\
\bottomrule
                         &              &               &  & &  &   \\
  
  \end{tabular}
}
  \caption{
         Summary of dataset tasks used for the tests.
  }
  \label{tab:dataset}
\end{table*}
%

%% file: figures/dataset_step_avg.tex
%
%
\begin{figure}[hb]
  \centering
  \includegraphics[width=1.0\columnwidth]{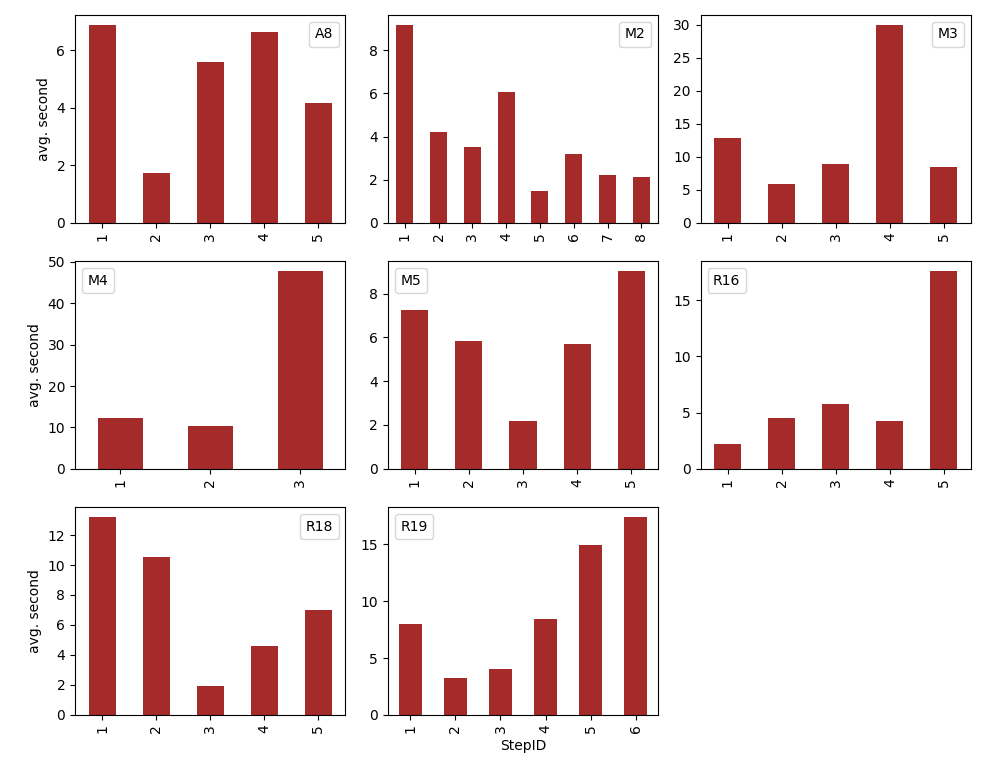}

  \caption{ Average step duration in seconds for the 8 selected tasks. }
  \label{fig:step_avg}
\end{figure}
%

%% file: s30_Dataset.tex
\section{The \datasetName~Dataset}
\label{sec:dataset}

\datasetName consists of \numTotalSkills tasks, summarized in the supplemental material, while this paper focuses on eight tasks that gained the most attention during the \ptg project.
Each task has a letter-number code, where the letter corresponds to the MARCH algorithm for treatment: massive bleeding, airway, respiration, circulation, head/hypothermia; and the number is a sequential identifier of tasks from each treatment.
Table~\ref{tab:dataset} presents the eight reported tasks with their codes and descriptions and a statistical summary with data recorded by professionals (professional data) and by our research team (in-lab data).
Examples of these distinct environments are presented in the left-most images of Figure~\ref{fig:summary}.
Furthermore, Figure~\ref{fig:step_avg} presents the average size of the steps, highlighting one of the data challenges: shorter steps can be more difficult to predict than longer ones.


\subsection{Data Collection}

\noindent\textbf{Professional Data.} To ensure that the professional data is properly represented, each task was performed by a trained Tactical Combat Casualty Care (TCCC) instructor.
Valkyries Austere Medical Solutions (VAMS)\footnote{https://valkyriesaustere.com/} was contracted to use their facilities, medical equipment, and trained instructors.
All data was collected with an IRB approved protocol\footnote{Approval from WCG IRB Solutions, WCGIRB Protocol Number: 20221136, HRPO Log Number E03430.1a.}.

Stereo images were collected with the Zed 2i by StereoLabs\footnote{https://www.stereolabs.com/store/products/zed-2i}, recording videos in 720x1280 resolution at 30 FPS.  
The dataset is distributed with .svo (Stereolabs proprietary format) and MP4 (from the left camera) files, which allows processing in 2D and/or 3D.
Since the Zed 2i cannot operate as a stand-alone sensor, it was connected to a Nvidia Jetson Nano\footnote{https://developer.nvidia.com/embedded/learn/get-started-jetson-nano-devkit} which was mounted on a military-style helmet, powered by an external battery pack.
Audio data was also collected by using a USB lavalier microphone clipped to the helmet near the Zed 2i.
It was recorded mono in WAV format (pcm\_s16le) with a 44.1KHz sample rate. 
The audio data is also available as a track in the corresponding MP4 file synchronized to the video (in which case it is transcoded to MP3 format, 44.1KHz, and 64kbps bit rate).

The Zed 2i helmet mount enables vertical adjustments, allowing the camera to move up and down.
All participants were instructed to collect an alignment image before recording using the software's alignment feature, and the camera was adjusted to ensure hands were close to the center of the alignment image.

This approach was used to record \numProVideos videos collected over 3 weeks on separate occasions.  
To add diversity, some parameters were varied, such as employing \numVAMSInstructors different instructors, 8 different locations, 13 volunteers, several different medical manikins, and many different medical equipment.
Figure~\ref{fig:summary} shows examples of possible locations in the left-most images and the A8 and M5 task examples.

\noindent\textbf{Additional Laboratory Data.}  
Additional videos were recorded to provide more controlled variations demanded by other \ptg teams, such as different participant expertise and camera field-of-view, along with environment changes, such as medical manikin position, light intensity, and background color.
Non-professional users recorded \numLabVideos videos with the Microsoft Hololens 2 (following \ptg needs) and referred to as ``in-lab dataset''.
The video resolution is 1920x1080 at 30 FPS with audio also available. 
Some examples can also be checked in Figure~\ref{fig:summary} left-most images and the M2 and R16 task examples.

\noindent\textbf{Evaluation Benchmark.}
In addition to the data presented above, 5 recordings were made for the \numSelectedSkills~selected tasks that are focused on. 
For each medical task, 4 out of the 5 videos were performed in the manner of an experienced individual, and one video was performed with hesitation and interruption designed to challenge the system. 
This evaluation data was collected on a Microsoft Hololens2 to match what the AR assistant was running on.
This configuration is used for the evaluations presented later in the text.

\subsection{Data Labeling}
There are three ways in which the data is labeled: object labels, action labels, and task-step labels (see Figure~\ref{fig:summary}).
For action and task steps, the labels are provided in frame numbers and timestamps.
Object labels are provided on a per-task basis in the YOLO label format.
Overall, there are bounding boxes for \numObjClasses classes, with a total of \numObjAnnos annotations.
Figure~\ref{fig:summary} presents bounding-boxes over the selected tasks in the right-most images. 

Action labels have the format $\texttt{time object action}$, where time is either a frame number or timestamp (in milliseconds).
Object describes what object is acting, such as the left hand or the right hand.
Action is the action being taken.
Overall, there are \numHandObjAnnos action label annotations on the \numProVideos professional videos (action labels were not created for the in-lab videos).
While the dataset provides this information, it is not used for the benchmark presented in this paper.

Task-step labels are similar to action labels and are used for the action detection benchmark presented in this paper.
They have the format $\texttt{start stop task\_step}$, where step is one of the predefined steps for the current task. 
A list of task steps is part of the data download for each task.
Overall, \numProSSAnnos task-steps were annotated in the \numProVideos professional videos, and \numLabSSAnnos task-steps were annotated in the \numLabVideos in-lab videos.
Figure~\ref{fig:summary} provides examples of still-images representing each task step of the most challenging tasks (A8, M2, M5, and R16), when it comes to number and duration of steps, shape-size of the main objects, steps overlap, or even view occlusion.

\noindent\textbf{Data and Evaluation Kit Availability.}
In total, the dataset comprises 786GB that will be available online and released under the CC BY-NC 4.0 license.
An evaluation kit, which was used to generate the results in this paper, will also be available at the same address.
The included documentation details the format and file structure of the kit and gives examples of input and output files. 


%% file: s40_Methodology.tex
\section{Action Detection Benchmark}
\label{sec:methodology}


Using the task step labels provided by the dataset, evaluation is presented as an action detection problem over which step is being performed.
This matches the larger objective of the \ptg program whereby assistance would be provided to a user unsure how to complete a medical task.  
Formally, this can be stated as, given a video $v$ from one task $k$ with a list of steps $S_k$, a model has to output a step $s\in S_k$ for each $v$-segment with $n$ frames. 

This work presents the benchmarking of three models created by \ptg performers.
The program structure had several independent teams developing algorithms for the AR medical assistant.  
The resulting AR medical system, called \magic, utilized an ensemble of these three approaches, whose discussion is outside the scope of this paper and may appear in a future publication.

\subsection{Recurrent and transformer approaches}
\label{sec:nyu_model}

\begin{figure*}[!h]
  \centering
   \includegraphics[scale=1]{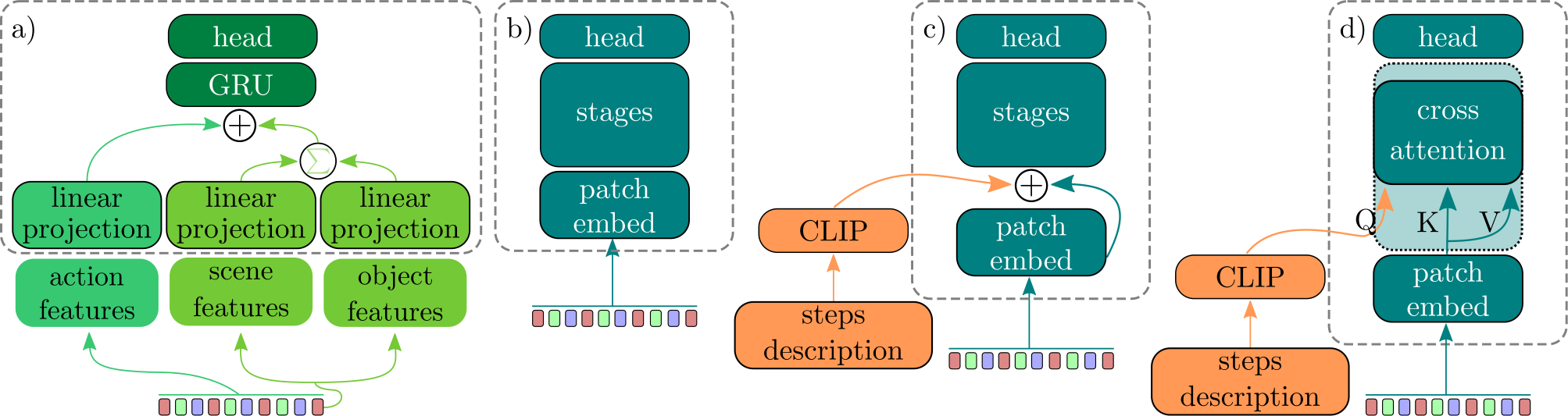} 
\caption{ RNN and transformer architectures: \textbf{a)} GRU-based model processes feature extracted from video window frames;
    and three variations of the Video Swin-transformer: 
    \textbf{b)} \emph{swin-basic}, fine-tuned to predict task steps;
    \textbf{c)} \emph{swin-concat}, which concatenates image tokens from video window frames and step description features;
    and \textbf{d)} \emph{swin-cross}, which replaces self-attention with cross-attention, using step description features as the query input. }
  \label{fig:nyu_models}   
\end{figure*}

To deal with challenging scenarios with different environmental conditions, object occlusion, and overlap of task steps, one of the \ptg research teams developed the following approaches. All code will be available online on GitHub.
Details about data preprocessing, training process, hyperparameters, and basic evaluation are presented in the supplemental material.

\subsubsection{Recurrent Neural Network (RNN)} 
In~\autoref{fig:nyu_models}a, the loading process slides a window over each video and extracts different feature modalities to feed a model.
As the task is based on action recognition, Omnivore~\cite{Girdhar:2022} is employed as an action-feature extractor (feature vector with 1024 positions) because it demonstrates strong generalization across different datasets, achieving high accuracy levels.

Following, for each window, a representative frame is identified (the last window frame in tests) and two sets of image-related features are extracted.
In the first set, called global (scene or frame) features, the CLIP~\cite{Radford:2021} image encoder is used to embed the frame into a feature space (512 positions), and in the second set the YOLO (v8)~\cite{Redmon:2016} models provided by \datasetName are used.
After extracting the detected objects from the frame, their image patches are embedded into a CLIP (region features) feature space (512 positions).
If the process fails to extract an object patch from the frame based on the YOLO detection, it returns a random image of the same dimensions.
The coordinates of the YOLO bounding box (normalized $xyxy$) and its confidence are concatenated at the end of each object feature vector (517 positions).
To avoid a variable number of object vector features, the process returns a zero-padded list of $y_c$ feature vectors, with $y_c$ being the number of classes YOLO was fine-tuned to predict.

In the sequel, the model projects global and region features with linear layers ($output = 512$, ReLU activation) and combines them using a weighted linear combination, with weights following the same idea of the ConceptFusion~\cite{Jatavallabhula:2023} process that calculates pixel-aligned features.
The process also averages the $y_c$ outputs of this combination to return one image representation per video window.
Finally, action and image features are projected with linear layers ($output = 512$, ReLU activation, and $dropout = 0.5$) and concatenated to feed two gated recurrent unit (GRU)~\cite{Cho:2014} layers followed by a ReLU activation and a linear layer that returns the final prediction. 
The recurrent characteristic of GRU helps to learn video spatial and temporal relationships to predict the steps. The input feature combination enables the use of different modalities.

The weights of all linear layers were initialized with the Xavier normal distribution~\cite{Glorot:2010}, the bias with zeros, and the GRU internal state with zeros in each training step.

\subsubsection{Transformer} The loading process slides a window over each video and uses the frames to feed a Video Swin-transformer~\cite{Liu:2022} to deal with the task-step identification.
This model presents suitable results on different video recognition benchmarks, including action recognition and temporal modeling, and three tests were performed with it.

In the first test, shown in~\autoref{fig:nyu_models}b (\emph{swin-basic}), the Swin-T version of the model was fine-tuned, pre-trained on Kinetics-400~\cite{Kay:2017}, only replacing the last linear layer with the right task-step number, initialized with a normal distribution ($mean = 0$ and $std = 0.02$ following the Torch implementation) and the bias with zeros.

In the second test in~\autoref{fig:nyu_models}c (\emph{swin-concat}), it was desired to leverage the step description semantics in the fine-tuning of the same Swin-T model version.
Simpler approaches were tested, different from ones such as BLIP-2~\cite{Li:2023}, which uses an intermediary model between a visual and text transformer.
The approach uses CLIP~\cite{Radford:2021} text encoder to embed the descriptions into a feature space (feature vector with 512 positions).
Furthermore, to match the same shape (\emph{batch size}, \emph{time}, \emph{hight}, \emph{width}, \emph{features}) of the original image tokens, the model \emph{i)} duplicates text samples to reach \emph{batch size}, \emph{ii)} and repeats the feature vectors to reach (\emph{time}, \emph{hight}, \emph{width}).
The model uses structures (layer normalization, linear layer with $output = 512$, and GELU activation) to learn projections of the image tokens and the text features.
The concatenation of these spaces (\emph{features} $=1024$ positions) is projected with a structure ($output = 96$) similar to the previous ones.
The concatenation output was also added with the initial image tokens in a skip connection to avoid convergence issues during backpropagation.
The result of these steps feeds the original model stages.
All added linear layers were also initialized with a normal distribution and the bias with zeros.

In the third test in~\autoref{fig:nyu_models}d (\emph{swin-cross}), it was desired to leverage the step description semantics in the Swin-T fine-tuning. 
However, the proposed model replaces all self-attention mechanisms with cross-attention, which uses the original model flow as the $KV$ vector and text embeddings as the $Q$ vector.
After loading the original attention parameters, the model splits $QKV$ joint weight/bias matrices to treat $Q$ ($1/3$ of the original matrices rows) separated from $KV$ ($2/3$ of the original matrices rows).
Similar to the previous test, it uses projection structures with outputs that match each transformer stage to learn projections of the text features used as $Q$ vectors.
All added linear layers were also initialized with a normal and the bias with zeros.

Therefore, there are three Video Swin-transformer variations but only the one with higher performance was selected for final evaluation: \emph{swin-basic} for M2, M3, M5, and A8 and \emph{swin-concat} for M4, R16, and R19.

\subsection{Temporal action segmentation approach}

\begin{figure}
    \centering
    \includegraphics[width=\linewidth]{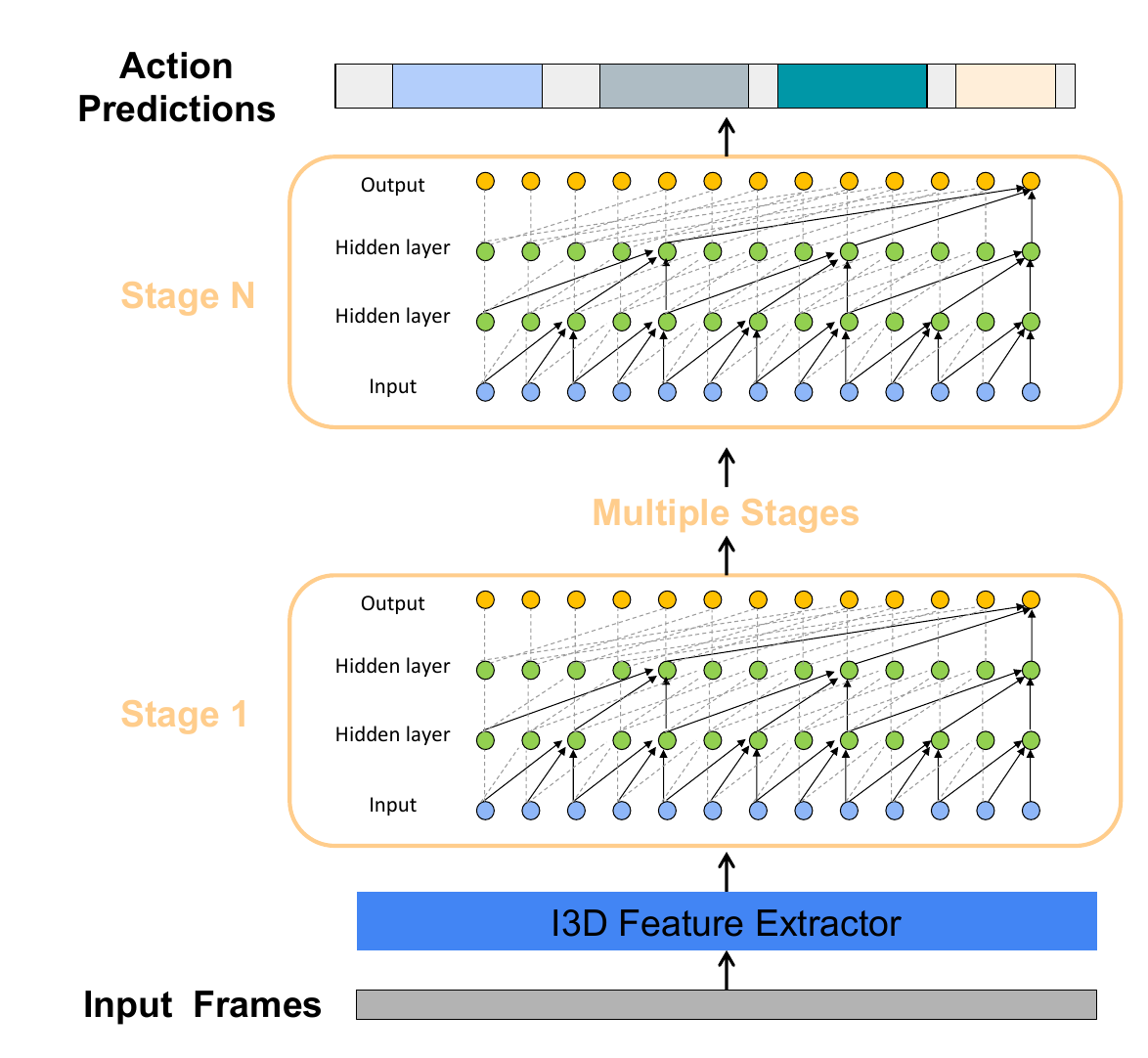}
    \caption{Temporal action segmentation architectures: a stack of multiple causal temporal convolution networks.}
    \label{fig:neu_framework}
\end{figure}

To deal with \datasetName scenarios, another PTG research team developed the following approach.

\subsubsection{Model Architecture}
This framework is based on a classical temporal action segmentation model, MSTCN~\cite{Farha:CVPR19}.
To enable online inference, the recent work ProTAS~\cite{Shen:CVPR24} was followed and the temporal convolutional network (TCN) modules were replaced with causal TCNs. 
This ensures that at any given time, the model only processes the frames observed up to that moment, thereby facilitating immediate action predictions. 
An overview of the model architecture is illustrated in Figure~\ref{fig:neu_framework}.

Concretely, I3D~\cite{Carreira:CVPR17} features are first extracted from incoming video frames using a sliding window of $32$ frames. 
The I3D feature extractor is pretrained on Kinetics-400~\cite{Carreira:CVPR17}, and outputs features of dimension $2048$ per frame. 
Let $f_t \in \mathbb{R}^{2048}$ denote the extracted feature for frame $t$. 
These features are used as input into a causal TCN module to obtain framewise action probabilities. 
Following MSTCN, a multi-stage refinement strategy is adopted. 
That is, the output action probabilities from one stage are fed into subsequent stages to progressively refine the predictions. 
Each stage consists of $10$ causal TCN layers, each with a hidden dimension of $64$. 
Four such stages are used in total.

Formally, let $p_{t,k}^s$ denote the probability of action class $k$ at frame $t$ and stage $s$, where $k \in \{1,2,\dots,K\}$ and $s \in \{1,2,\dots,N\}$. 
Here, $K$ is the number of action classes and $N$ is the number of stages. 
The first stage takes as input the I3D features $f_t$, and the output of each subsequent stage $s>1$ refines the probabilities from stage $s-1$.

\subsubsection{Training}
A separate model is trained for each of the eight medical tasks. 
Each task is associated with a set of action classes that define a canonical sequence in which these actions typically appear. 
Let $\{y_t\}_{t=1}^T$ be the ground-truth action labels for a training video of length $T$ frames, where $y_t \in \{1,2,\dots,K\}$.

A loss function is employed that encourages correct classification while also enforcing temporal smoothness. Specifically, the loss $L$ at training stage $s$ is a combination of a cross-entropy loss and a smoothing loss. 
The cross-entropy loss ensures that the predicted probabilities $p_{t,y_t}^s$ are maximized for the correct actions, and the smoothing loss penalizes large temporal fluctuations between consecutive frames. 
Formally:
\begin{equation}
    L = \sum_{s=1}^{N} \sum_{t=1}^{T} \biggl[-\log(p_{t,y_t}^s) + \lambda \sum_{k=1}^K (p_{t,k}^s - p_{t-1,k}^s)^2 \biggr],
\end{equation}
where $p_{t-1,k}^s$ is considered as $0$ for $t=1$. 
Here, the first term $-\log(p_{t,y_t}^s)$ maximizes the log-likelihood of the correct action class, and the second term $(p_{t,k}^s - p_{t-1,k}^s)^2$ encourages temporal smoothness by penalizing abrupt changes in the predicted probabilities between consecutive frames. 
The hyperparameter $\lambda$ controls the relative weight of the smoothing term.

During training, the videos and associated labels are randomly trimmed and the models are trained on these trimmed videos to enhance the model's robustness towards limited data.

\subsubsection{Inference}
Inference is performed in an online manner. 
At any given time, all frames received up to that moment are accessible. 
I3D features are extracted on-the-fly from incoming frames using a sliding window of size $32$. 
To manage memory and computation costs, a buffer of the most recent $1200$ frames is maintained and older frames are discarded once the buffer limit is reached.

The buffered features are fed into the four-stage causal TCN model to obtain framewise action probabilities at the current time step. 
To ensure that the final predictions respect the canonical ordering of actions for the given task, an additional refinement is applied to the logits before converting probabilities into discrete action labels.

Let $z_{t,k}$ denote the raw logit for action $k$ at time $t$ after the final refinement stage. 
Suppose that a predefined sequence of actions $(a_1, a_2, \dots, a_K)$  defines the typical temporal progression of the task. 
For each action $a_k$, let $P(a_k)$ and $S(a_k)$ denote the set of its predecessors and successors in the default sequence. 
During inference, the actions that have been predicted so far are tracked.
If an action $a_k$ is being considered at time $t$ but some of its predecessors have not yet been completed, a penalty on $z_{t,k}$ is imposed.
Similarly, a penalty will be imposed if some of its successors have already been detected. 
Formally, let $B_{a}(t)$ be an indicator variable that equals $1$ if an action $a$ has been completed by time $t$, and $0$ otherwise.
The logits are modified as follows:
\begin{equation}
    \tilde{z}_{t,k} = z_{t,k} - \alpha \left(\sum_{a \in P(a_k)} (1 - B_a(t)) + \sum_{a \in S(a_k)}B_a(t) \right),
\end{equation}
where $\alpha > 0$ is a fixed penalty coefficient. 
This adjustment ensures that actions are more likely to appear in the correct temporal order, discouraging the model from prematurely predicting later actions whose predecessors have not yet been observed or predicting earlier actions whose successors have already been observed. 
Finally, a softmax is applied into the modified logits to compute the probabilities for each action class.

%% file: s50_Results.tex
\section{Experimental Results}
\label{sec:results}

\begin{table*}[hbt]
  \centering
  \begin{tabular}{ lcccccc }
    \toprule
    Method & mAP@0.1	&	mAP@0.2	&	mAP@0.3	&	mAP@0.4	&	mAP@0.5	&	Avg mAP \\
    \midrule
TAS	    &	0.437	&	0.372	&	0.271	&	0.197	&	0.129	&	0.281 \\
RNN	    &	0.732	&	0.652	&	0.504	&	0.402	&	0.339	&	0.526 \\
Swin-T	&	0.732	&	0.615	&	0.523	&	0.396	&	0.301	&	0.513 \\
    \bottomrule
  \end{tabular}
  \caption{Mean Average Precision at different IOU thresholds across the benchmark methods: Temporal action segmentation (TAS), RNN, and Swin-T.}
  \label{tab:results}
\end{table*}

\begin{table*}[hbt]
  \centering
  \resizebox{\textwidth}{!}{
  \begin{tabular}{ lcccccccccccccccc }
    \toprule
Method & \multicolumn{2}{c}{A8} & \multicolumn{2}{c}{M2} & \multicolumn{2}{c}{M3} & \multicolumn{2}{c}{M4} & \multicolumn{2}{c}{M5} & \multicolumn{2}{c}{R16} & \multicolumn{2}{c}{R18} & \multicolumn{2}{c}{R19} \\
\toprule
&Prec&Rec&Prec&Rec&Prec&Rec&Prec&Rec&Prec&Rec&Prec&Rec&Prec&Rec&Prec&Rec \\
\cmidrule(lr){2-3}
\cmidrule(lr){4-5}
\cmidrule(lr){6-7}
\cmidrule(lr){8-9}
\cmidrule(lr){10-11}
\cmidrule(lr){12-13}
\cmidrule(lr){14-15}
\cmidrule(lr){16-17}
TAS&0.343&0.48&0.308&0.343&0.25&0.28&0.312&0.357&0.5&0.292&0.4&0.32&0.5&0.458&0.333&0.241 \\
RNN&0.8&0.64&0.421&0.216&0.909&0.4&0.909&0.667&0.5&0.12&1&0.28&0.846&0.458&0.875&0.467 \\
Swin-T&0.824&0.56&0.5&0.27&0.727&0.32&0.909&0.667&0.75&0.24&0.8&0.32&0.714&0.417&0.714&0.333 \\
    \bottomrule 
  \end{tabular}
  }
  \caption{Precision and Recall values across all 8 tasks with an IOU threshold of 0.5.}
  \label{tab:results_pr}
\end{table*}

As previously stated, videos are labeled for the start and stop time for each step in the task.
These task step intervals are the activity segments for the action detection problem.
The segments are compared with ground truth to compute an intersection-over-union (IOU) score.
If the IOU is equal to or greater than a threshold, then there is a match between the prediction and the ground truth.
Varying the threshold allows the calculation of precision-recall curves, along with the average precision (AP) summarizing performance for each task.
The resulting AP's can then be averaged across tasks to compute a mean average precision (mAP) for each IOU threshold.
All of these computations are performed with the released evaluation kit\footnote{\datasetUrl}.
We refer to the models described in Section~\ref{sec:methodology} as Temporal action segmentation (TAS), RNN, and Swin-T, independently of the transformer variation tested with the task. 
Lastly, we named all tasks with their codes listed in Table~\ref{tab:dataset}.

Table~\ref{tab:results} presents mAP scores at IOU thresholds from 0.1 to 0.5 and an average mAP for each benchmark model, over the eight tested tasks.
In addition, Table~\ref{tab:results_pr} shows precision and recall values for each method, across the tested tasks at an IOU threshold of 0.5.

Overall, TAS performs worse than RNN or Swin-T. However, across most tasks, its precision and recall scores remain relatively close, indicating a balanced performance. M5 and R18 achieve the highest precision (0.5), suggesting the effectiveness at minimizing \emph{false positives}, while M3 has the lowest precision (0.25), implying a higher tendency for incorrect predictions. In terms of recall, A8 performs the best (0.48), capturing the most relevant instances, whereas R19 has the lowest recall (0.241), indicating difficulty in retrieving all relevant instances. These results highlight that, while TAS maintains a trade-off between precision and recall, it still falls short compared to more advanced sequence modeling approaches.

RNN and Swin-T achieved the best results on M4, with precision up to 0.91 and recall up to 0.67.
On the other hand, the worst results were obtained on M2, with precision up to 0.50 and recall up to 0.27.
This scenario is influenced by two factors: 1) M4 and M2 consist of three and eight steps, respectively, whereas most of the other evaluated tasks contain five steps (see Table~\ref{tab:dataset}); and 2) the longest M4 step involves an object with minimal attribute changes. 
In contrast, the tourniquet is used throughout all M2 steps, undergoing multiple physical changes, which impair object and step identification. Ultimately, tourniquets are nylon straps with a tightening mechanism and highly deformable.
They are often looped, meaning that the center of their bounding box will be more in the background than in the foreground.
Additionally, at least two M2 steps are absent in part of the M2 videos.

Following, it can be observed that RNN and Swin-T present significant differences between precision (values $\in [0.42, 1.00]$) and recall (values $\in [0.12, 0.67]$).
This indicates that models generate fewer \emph{false positives}, improving precision, but produce more \emph{false negatives}, which lowers recall.
The high \emph{false negatives} rate can be attributed to 1) overlap between the steps, as seen in the M2 task; 2) confusion in identifying objects due to subtle movements of camera or user hands; and 3) variations in step duration (see Figure~\ref{fig:step_avg}) and user expertise.

These issues pose challenges despite considering well-suited techniques such as Omnivore, YOLO, and GRU for detecting user actions and objects across different datasets, along with action and objects long-term dependencies.
Furthermore, hierarchical representation in video transformers, with varying image and time patch sizes, helps identifying objects with different shapes and sizes.
This structure and the attention mechanisms capture long-range dependencies, facilitating the identification of correlated steps and actions.

While offline processing could improve action detection performance, \ptg requires real-time AR system usage.
Besides, Table~\ref{tab:runtimes} lists the run times for the models' inference.
Inference for TAS was run on one Nvidia RTX 6000 GPU with a Nvidia driver v515.65.01.
Inference for RNN and Swin-T was run on the one Nvidia A100 GPU per model and Nvidia driver v535.154.05.

\begin{table}[h]
  \centering
  \begin{tabular}{ lc }
    \toprule
    Method & Inference Time (ms) \\
    \midrule
TAS	    &	510 \\
RNN	    &	210 \\
Swin-T	&	330 \\
    \bottomrule
  \end{tabular}
  \caption{Average inference time across the benchmark methods: Temporal action segmentation (TAS), RNN, and Swin-T.}
  \label{tab:runtimes}
\end{table}

%% file: s60_Conclusions.tex
\section{Conclusions \& Future Work}
\label{sec:conclusion}

This paper presents \datasetName, an egocentric dataset that contains \numTotalVideos videos of tactical combat casualty care tasks.
Each task has more than 50 video examples and was labeled for action recognition, action detection, and object detection.
A subset of eight of the original 50 tasks was selected for more in-depth analysis on action detection using three different approaches, the best of which achieved an average mAP score of 0.526.
The results highlight the potential of \datasetName for presenting challenges for the computer vision community, related to adaptation and generalization of the techniques used for action identification and detection.

Key directions for future work include: (1) establishing a comprehensive performance benchmark that leverages all 50 recorded medical tasks, (2) exploring the integration of stereo vision and audio data to enhance recognition capabilities and comparative analyses, and (3) investigating the impact of domain shifts on model performance. The latter could involve experimenting with different data modalities—such as stereo vs. RGB-only input—or assessing variations in expertise levels (amateur vs. professional) and hardware configurations (e.g., HoloLens vs. Zed), which differ significantly in their fields of view.

Additionally, future work could focus on advancing medical task recognition by analyzing sequences of multiple tasks performed in succession or concurrently across multiple casualties. The dataset also includes a limited collection of annotated error recordings, which could serve as the foundation for research challenges, such as error detection and automated assessment of procedural deviations. These directions have the potential to further the capabilities of AI-driven medical training and performance evaluation, making the \datasetName a valuable resource for continued exploration in computer vision and healthcare applications.


%% file: s99_References.bib
@article{vanvoorst2023automated,
  title={Automated Video Debriefing Using Computer Vision Techniques},
  author={Brian VanVoorst et al.},
  journal={Simulation in Healthcare},
  volume={18},
  number={5},
  pages={326--332},
  year={2023},
  publisher={LWW}
}

@inproceedings{walczak2022coach,
  title        = {Automating Video After Action Reviews for Military Medical Training},
  author       = {Nicholas R Walczak et al.},
  year         = 2022,
  month        = {November},
  booktitle    = {Proceedings of Interservice/Industry Training, Simulation and Education Conference (I/ITSEC)},
}

@InProceedings{thompsontrauma,
author="Yupeng Zhuo et al.",
title="Overview of the Trauma THOMPSON Challenge at MICCAI 2023",
booktitle="AI for Brain Lesion Detection and Trauma Video Action Recognition",
year="2025",
publisher="Springer Nature Switzerland",
address="Cham",
pages="47--60",
isbn="978-3-031-71626-3"
}

@misc{epickitchens,
      title={The EPIC-KITCHENS Dataset: Collection, Challenges and Baselines}, 
      author={Dima Damen et al.},
      year={2020},
      eprint={2005.00343},
      archivePrefix={arXiv},
      primaryClass={cs.CV},
      url={https://arxiv.org/abs/2005.00343}, 
}

@INPROCEEDINGS{Girdhar:2022,
    author    = {Rohit Girdhar et al.},
    title     = {{Omnivore: A Single Model for Many Visual Modalities}},
    booktitle = {{Proceedings of the IEEE/CVF Conference on Computer Vision and Pattern Recognition (CVPR)}},
    month     = {June},
    year      = {2022},
    pages     = {16102-16112}
}

@INPROCEEDINGS{Radford:2021,
  author={Alec Radford et al.},
  title={Learning Transferable Visual Models From Natural Language Supervision},
  booktitle={Proceedings of the 38th International Conference on Machine Learning},
  pages={8748--8763},
  year={2021},
  volume={139},
  series={Proceedings of Machine Learning Research},
  month={18--24 Jul},
  publisher={PMLR}
}

@INPROCEEDINGS{Redmon:2016,
  author = {Joseph Redmon et al.},
  title = {You Only Look Once: Unified, Real-Time Object Detection},
  booktitle = {Proceedings of the IEEE Conference on Computer Vision and Pattern Recognition (CVPR)},
  month = {June},
  year = {2016}
}

@ARTICLE{Jatavallabhula:2023,
  author    = {{Krishna Murthy} Jatavallabhula et al.},
  title     = {ConceptFusion: Open-set Multimodal 3D Mapping},
  journal   = {arXiv},
  year      = {2023},
}

@INPROCEEDINGS{Glorot:2010,
  author = 	 {Glorot, Xavier and Bengio, Yoshua},
  title = 	 {Understanding the difficulty of training deep feedforward neural networks},
  booktitle = 	 {Proceedings of the Thirteenth International Conference on Artificial Intelligence and Statistics},
  pages = 	 {249--256},
  year = 	 {2010},
  editor = 	 {Teh, Yee Whye and Titterington, Mike},
  volume = 	 {9},
  series = 	 {Proceedings of Machine Learning Research},
  month = 	 {13--15 May},
  publisher =    {PMLR},
}

@INPROCEEDINGS{Liu:2022,
    author    = {Ze Liu et a.},
    title     = {Video Swin Transformer},
    booktitle = {Proceedings of the IEEE/CVF Conference on Computer Vision and Pattern Recognition (CVPR)},
    month     = {June},
    year      = {2022},
    pages     = {3202-3211}
}

@MISC{Kay:2017,
    author={Will Kay et al.},
    title={The Kinetics Human Action Video Dataset}, 
    year={2017},
    eprint={1705.06950},
    archivePrefix={arXiv},
    primaryClass={cs.CV},
    url={https://arxiv.org/abs/1705.06950}, 
}

@MISC{Cho:2014,
      author={Kyunghyun Cho et al.},
      title={Learning Phrase Representations using RNN Encoder-Decoder for Statistical Machine Translation}, 
      year={2014},
      eprint={1406.1078},
      archivePrefix={arXiv},
      primaryClass={cs.CL},
      url={https://arxiv.org/abs/1406.1078}, 
}

@INPRODEEDINGS{Gorpincenko:2023,
    author={Gorpincenko, Artjoms and Mackiewicz, Michal},
    title={Extending Temporal Data Augmentation for Video Action Recognition},
    booktitle={Image and Vision Computing},
    year={2023},
    publisher={Springer Nature Switzerland},
    pages={104--118}
}

@article{Shen:CVPR24,
  author = {Y.~Shen and E.~Elhamifar},
  title = {Progress-Aware Online Action Segmentation for Egocentric Procedural Task Videos},
  journal = {{IEEE} Conference on Computer Vision and Pattern Recognition},
  year = {2024}}

@inproceedings{Farha:CVPR19,
  title={Ms-tcn: Multi-stage temporal convolutional network for action segmentation},
  author={Farha, Yazan Abu and Gall, Jurgen},
  booktitle={Proceedings of the IEEE/CVF conference on computer vision and pattern recognition},
  pages={3575--3584},
  year={2019}
}

@conference{Carreira:CVPR17,
   Author ={J.~Carreira and A.~Zisserman},
   Title = {Quo Vadis, Action Recognition? A New Model and the Kinetics Dataset},
   Booktitle = {IEEE Conference on Computer Vision and Pattern Recognition},
   Year = {2017}}

@MISC{Li:2023,
    author={Junnan Li et al.},
    title={BLIP-2: Bootstrapping Language-Image Pre-training with Frozen Image Encoders and Large Language Models}, 
    year={2023},
    eprint={2301.12597},
    archivePrefix={arXiv},
    primaryClass={cs.CV}
}

@BOOK{Goodfellow:2016,
  author={Goodfellow, Ian and Bengio, Yoshua and Courville, Aaron},
  title={Deep learning},
  year={2016},
  publisher={MIT press}
}

@ARTICLE{Vaswani:2017,
  author={Ashish Vaswani et al.},
  title={Attention is all you need},
  journal={arXiv preprint arXiv:1706.03762},
  year={2017}
}

@INPROCEEDINGS{Goyal:2017,
  author = {Raghav Goyal et al.},
  title = {The ``Something Something'' Video Database for Learning and Evaluating Visual Common Sense},
  booktitle = {Proceedings of the IEEE International Conference on Computer Vision (ICCV)},
  month = {Oct},
  year = {2017}
}

@INPROCEEDINGS{Grauman:2022,
    author    = {Kristen Grauman et al.},
    title     = {j},
    booktitle = {Proceedings of the IEEE/CVF Conference on Computer Vision and Pattern Recognition (CVPR)},
    month     = {June},
    year      = {2022},
    pages     = {18995-19012}
}

@ARTICLE{Haiha:2016,
  author       = {Sami Abu{-}El{-}Haija et al.},
  title        = {YouTube-8M: {A} Large-Scale Video Classification Benchmark},
  year         = {2016},
  eprint       = {1609.08675},
  archivePrefix= {arXiv},
  journal      = {CoRR}
}

@MISC{PTG_site,
  author = {{DARPA}},
  title = {Perceptually-enabled Task Guidance {(PTG)}},
  howpublished = "\url{https://www.darpa.mil/program/perceptually-enabled-task-guidance}"
}

@INPROCEEDINGS{Fathi:2011,
  author={Fathi, Alireza and Ren, Xiaofeng and Rehg, James M.},
  booktitle={CVPR 2011}, 
  title={Learning to recognize objects in egocentric activities}, 
  year={2011},
  volume={},
  number={},
  pages={3281-3288}
}

@INPROCEEDINGS{Li:2015,
    author = {Li, Yin and Ye, Zhefan and Rehg, James M.},
    title = {Delving Into Egocentric Actions},
    booktitle = {Proceedings of the IEEE Conference on Computer Vision and Pattern Recognition (CVPR)},
    month = {June},
    year = {2015}
}

@ARTICLE{Pan:2022,
  author={Junhao Pan et al.},
  title={YouHome System and Dataset: Making Your Home Know You Better},
  journal={IEEE International Symposium on Smart Electronic Systems (IEEE - iSES)},
  year={2022}
}

@book{phtls2019,
  author    = {{National Association of Emergency Medical Technicians (NAEMT)}},
  title     = {PHTLS: Prehospital Trauma Life Support},
  edition   = {9th},
  publisher = {Jones \& Bartlett Learning},
  address   = {Burlington, MA},
  year      = {2019},
  isbn      = {9781284180586}
}
